\documentclass{article} 
\usepackage{nips13submit_e,times}
\usepackage{url}
\usepackage{graphicx}

\title{Sparse Code Formation with Linear Inhibition}

\author{
Nam Do-Hoang Le\\
John von Neumann Institute, VNU-HCMC\\
Ho Chi Minh city\\
\texttt{nam.le.ict@jvn.edu.vn} \\
}

\nipsfinalcopy 

\begin{document}

\maketitle

\section{Introduction}
Sparse code formation in the primary visual cortex (V1) has been inspiration for many state-of-the-art visual recognition systems. To stimulate this behavior, networks are trained networks under mathematical constraint of sparsity or selectivity. Meanwhile, there is another line of research which emphasizes the role of lateral connections of neural networks in sparse code formation. Lateral connections are synapses among neurons on the same layer, which is an essential part of human neural networks. There are two types of interconnections. Excitatory connections propagate firing signal across neural layer, thus preserve topographical order of neural stimuli. On the other hand, inhibitory connections decorrelate activations among neurons, which accounts for sparse code formation.

The key challenge when implementing interconnected networks is the high complexity of the structure. Firing value of one neuron is also input for other neurons on the same layer through lateral synapses. Spike output of each neuron is determined by the membrane potential, which is accumulated over an amount of time \cite{zylberberg2011, king2013}. In other words, a few iterations are needed for each layer to stabilize the firing rates. Thus it is challenged to scale up networks for large-scaled learning with enormous amount of unlabeled data. To tackle this problem, the authors reach to an inspiration from Dale's law \cite{eccles1976} which observes that one cortical neuron is either inhibitory or excitatory, but not both \cite{king2013}. Inspired by Dale's law , we construct an unsupervised neural network which enables sparse learning with lateral inhibition but still practically trainable. The key component of our networks is the inhibitory layer placed after the normal hidden layer (i.e. encoding layer). Each neuron in this inhibitory layer decreases the firing rate of one corresponding hidden neuron by an linear combination of other neural firing rates. The more correlated two neurons are, the stronger the inhibitory weight between them is. Hence, the final features are more robust due to omitting redundant correlated information among neurons. To learn the inhibitory weights, we employ Hebbian learning rule, a simple plasticity learning rule. Neurons on the first hidden layer are trained normally with any unsupervised algorithms.

\section{Inhibitory layer in networks}
\label{in_layer}
\begin{figure}[h]
\label{figure1}
\begin{center}
\includegraphics{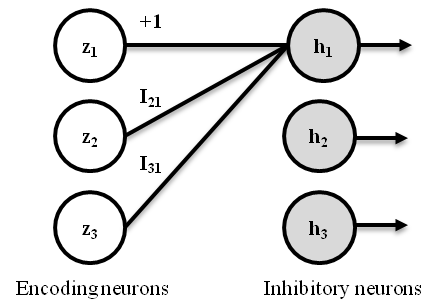}
\end{center}
\caption{A toy example of encoding - inhibitory layer. In which, $h_1 = z_1 - (I_{21}z_2 + I_{31}z_3)$}
\end{figure}

In this section, we present the core improvement of our system, the inhibitory layer. This layer is placed right after the encoding layer to refine the feature representation. Figure \ref{figure1} illustrates a toy example of this structure.

Following the basic idea of inhibition, each neuron is suppressed directly by firing rates of other neurons on the same layer. To reduce complexity from accumulation on membranes, we preserve the feed-forward structure by assign a new layer of specific neurons for inhibition. Each signal is weighted by an inhibitory factor on the connection to corresponding neuron. All encoding signals from encoding layer are integrated to suppress firing rates. Concretely, the process could be described as following: 

\begin{itemize}
\item Let $z_i$ $(i = 1..K)$ be firing rate of hidden neuron $i$. There are $K$ corresponding inhibitory neurons with the output is $h_i$. Encoding neurons are fully connected to inhibitory neurons.
\item For $i, j = 1..K$, $I_{ji}$ is the weight of connection from encoding neuron $j$ to inhibitory neuron $i$.
\item Let $h_i$ be the output of the corresponding $z_i$. $h_i$ is determined by $z_i$ surpressed by a linear combination of $z_j$ and $I_{ji}$, for all $j \neq i$:
\[h_i = max(0, (z_i - \sum_{j \neq i}{I_{ji}z_j}))\]
\end{itemize}

At first, $I_{ji}$ is initialized uniformly, which makes inhibitory layer behave similarly to triangle K-Means \cite{coates2011}. Then $I_{ji}$ must be trained such that the more correlated two neurons are, the stronger their inhibition is. To achieve this objective, Hebbian rule is employed on the inhibitory layer. Lateral weights are also normalized to keep the sum of weights constant. For every $j \neq i$:
\[I_{ji}^{(new)} = \frac{I_{ji}^{(old)} + \alpha z_{i}h_{j}}{Z_i}\]
\[Z_i = \sum_{j \neq i}{I_{ji}^{(new)}}\]

In our choice of encoding layer, there are no excitatory connections, encoding neurons are not located in topographical order. Without order, a neighborhood of one neuron cannot be determined. Therefore, one neuron is inhibited by all other encoding neurons. To create the notion of neighborhood, after some epochs, weak inhibitory links are omitted. One neuron is now inhibited by only a small number of neurons in the neighborhood. This number of neighbors can be chosen as hyper-parameter or can be determined by omitting weak links after each epoch.

\section{Experiments}
\label{exre}

To verify effects of inhibitory layer, we conducted experiments in a standard dataset for visual recognition problem: CIFAR-10 \cite{krizhevsky2009}. In all cases, one image is processed using general framework with a partition of 4 pooling regions. Each region is pooled using average pooling. Linear classifier is softmax regression with parameters determined through cross-validation. Vector quantization and sparse autoencoder \cite{hinton2006} are chosen to train encoding layer.

Table \ref{cifar-table1} shows the improvement in comparison with standard approach using sparse autoencoder. Without determining neighborhood, one neuron is inhibited by all other neurons in encoding layer. The margin widens as the number of features increases. We can temporarily explain this phenomenon as when the number of features increases, one neuron has more correlated neurons and it is more likely to have noise among neurons.
\begin{table}[t]
\caption{Improvement in sparse autoencoder}
\label{cifar-table1}
\begin{center}
\begin{tabular}{lll}
\multicolumn{1}{c}{\bf \#Hidden}  &\multicolumn{1}{c}{\bf Sparse AE} &\multicolumn{1}{c}{\bf Inhibitory AE}
\\ \hline \\
400         &72.95  	&73.42 \\
800         &73.86 		&75.17 \\
1200        &73.63  	&75.86 \\
1600				&73.96 		&76.02 \\
\end{tabular}
\end{center}
\end{table}

Because features are learnt independently with inhibitory layer, inhibitory networks are still considered shallow. In second experiment, inhibitory layer is added to K-Means with the size of neighborhood is 40. Inhibitory K-Means shows promising result when being compared with other state-of-the-art single-layer methods in table \ref{cifar-table2}. Inhibitory AE and K-Means outperform conventional unsupervised algorithms. However, there is a gap from our method to another augmented networks, sparse TIRBM.
\begin{table}[t]
\caption{Comparison with other single-layer networks ( $^1$reported results)}
\label{cifar-table2}
\begin{center}
\begin{tabular}{ll}
\multicolumn{1}{c}{\bf Methods}  &\multicolumn{1}{c}{\bf Accuracy}
\\ \hline \\
Sparse RBM$^1$ \cite{coates2011}			&72.4 \\ 
Sparse AE													&73.96 \\
Improved LCC$^1$ \cite{yu2010}				&74.5 \\
Triangle K-Means \cite{coates2011}	&77.68 \\
\textbf{Inhibitory K-Means}	&77.95 \\
Sparse TIRBM$^1$ \cite{sohn2012}			&78.8 \\
\end{tabular}
\end{center}
\end{table}
\section{Related Works}
Lateral connections have been widely used in models in neuroscience. Our work is loosely based on E-I net \cite{king2013}. In E-I net, simple cells are divided into two types of neurons, excitatory neurons and inhibitory neurons. One inhibitory cell sends an amount of inhibitory current directly to all excitotary simple cells and other inhibitory cells. Inhibitory cells predict the redundant part of the network activity, thus decorrelate the activity of the excitatory cells by suppressing redundant spiking activity. In our networks, relationship between encoding cells is stored in connections between two layers. Inhibitory cells only do the computation based on encoding signals and the corresponding weights. In one recent work that shares the idea of local computation \cite{srivastava2013}, spiking signal is chosen using winter-take-all rule while the rest is grounded to 0. Moreover, the competition range is predefined. In other works aiming at learning structural features such as structural sparse coding \cite{szlam2011} and topographical ICA \cite{le2011}, structural order emerges through regularization. Meanwhile, with inhibitory layer, neighborhood rather than order is obtained through ad hoc optimization.
\section{Conclusion}
In this paper, the authors propose an idea of adding inhibitory layer into feature learning networks to boost the performance. Using sparse autoencoder and K-Means as basic encoding layer, we have demonstrated the properties of inhibitory autoencoder. A result on CIFAR-10 shows potential improvement on visual recognition. However, it also leaves open questions for further research such as whether inhibitory layer contributes into the emergence of features on the encoding layer. Sparse code created by inhibitory layer could play as the desired output for encoding layer in the learning process.

\bibliographystyle{ieeetr}
\bibliography{nips2013}
\end{document}